\documentclass{article}
\usepackage[preprint,nonatbib]{neurips_2025}
\usepackage{amsmath, amssymb}
\usepackage{graphicx}
\usepackage{authblk}
\usepackage[utf8]{inputenc} % allow utf-8 input
\usepackage[T1]{fontenc}    % use 8-bit T1 fonts
\usepackage{hyperref}       % hyperlinks
\usepackage{url}            % simple URL typesetting
\usepackage{booktabs}       % professional-quality tables
\usepackage{amsfonts}       % blackboard math symbols
\usepackage{nicefrac}       % compact symbols for 1/2, etc.
\usepackage{microtype}      % microtypography
\usepackage{xcolor}         % colors

\title{Directional Non-Commutative Monoidal Structures for Compositional 
Embeddings in Machine Learning\thanks{We thank Prof. S. Prem Kumar at Swansea University for their valuable insights. A patent application
has been filed for aspects of this work.}}

\author[1]{Mahesh Godavarti}
\affil[1]{Qalaxia \\
\texttt{m@qalaxia.com}}

\begin{document}

\maketitle

\begin{abstract}
We introduce a new algebraic structure for multi-dimensional compositional
embeddings, built on directional non-commutative monoidal operators. The core
contribution of this work is this novel framework, which exhibits appealing
theoretical properties (associativity along each dimension and an interchange
law ensuring global consistency) while remaining compatible with modern machine
learning architectures. Our construction defines a distinct composition operator
$\circ_i$ for each axis $i$, ensuring associative combination along each axis
without imposing global commutativity. Importantly, all axis-specific operators
commute with one another, enforcing a global interchange law that enables
consistent cross-axis compositions. This is, to our knowledge, the first approach
that provides a common foundation that generalizes classical sequence-modeling paradigms 
(e.g., structured state-space models (SSMs) and transformer self-attention) to a unified multi-dimensional framework. 
For example, specific one-dimensional instances of our framework can recover the familiar affine transformation algebra,
vanilla self-attention, and the SSM-style recurrence. The higher-dimensional generalizations naturally support 
recursive, structure-aware operations in embedding spaces.
We outline several potential applications unlocked by this structure---including structured positional
encodings in Transformers, directional image embeddings, and symbolic modeling
of sequences or grids---indicating that it could inform future deep learning
model designs. We formally establish the algebraic properties of our framework
and discuss efficient implementations. Finally, as our focus is theoretical, we
include no experiments here and defer empirical validation to future work, which
we plan to undertake.
\end{abstract}

\section{Introduction}

Many forms of structured data can be composed hierarchically along one or more dimensions. In the one-dimensional (1D) case, a sequence like $[a\ b\ c\ d]$ can be formed by composing sub-sequences—for example, by concatenating $[a\ b]$ with $[c\ d]$, or $[a]$ with $[b\ c\ d]$. Algebraic tools such as non-commutative semigroups or free groups have long provided principled ways to model such 1D compositional structure \cite{Rudolph2010}.

In contrast, there exists no broadly accepted algebraic framework for modeling \emph{two-dimensional (2D) composition}. Consider the array:
\[
\begin{bmatrix}
a & b \\
c & d
\end{bmatrix}
\]
This structure can be composed either vertically, by stacking
\[
\begin{bmatrix}
a & b
\end{bmatrix}
\quad \text{over} \quad
\begin{bmatrix}
c & d
\end{bmatrix},
\]
or horizontally, by placing
\[
\begin{bmatrix}
a \\
c
\end{bmatrix}
\quad \text{next to} \quad
\begin{bmatrix}
b \\
d
\end{bmatrix}.
\]

Such multiple valid composition paths do not fit neatly into existing algebraic systems, which are typically designed for linear (1D) structured data. Even in the 1D case, aligning algebraic formalisms with the needs of modern machine learning architectures can be problematic—for example, representing tokens as matrices instead of vectors breaks the core assumptions of the attention mechanism in transformers, which relies fundamentally on vector operations.

This tension reveals a deeper issue: the absence of an algebraic framework that can both align with vector-based learning architectures and naturally support composition along multiple axes. 

The lack of such a model limits our ability to represent and reason about grid-structured data in a principled way. This is a pressing issue in machine learning domains like natural language processing and computer vision, where capturing rich structural relationships is key to performance. Yet, despite advances in encoding positional and relational information, current methods lack a rigorous algebraic underpinning for multi-dimensional, direction-sensitive composition.

Therefore, what is needed is a new algebraic structure—one that unifies 1D and higher-dimensional composition under a coherent, machine-compatible foundation.

We present such a structure defined by:
\begin{itemize}
    \item A distinct composition operator $\circ_i$ for each axis $i$,
    \item Associativity along each axis: $(x \circ_i y) \circ_i z = x \circ_i (y \circ_i z)$,
    \item A global interchange law holding across all axes: $(x \circ_i y) \circ_j (u \circ_i z) = (x \circ_j u) \circ_i (y \circ_j z)$,
    \item Non-commutativity of the individual axis-specific composition operators: $x \circ_i y \neq y \circ_i x$.
\end{itemize}

To our knowledge, we are the first to present such a formulation.

For clarity, this new algebraic construction is the central contribution of our work, offering both a rigorous theoretical foundation and promising practical implications. By addressing the aforementioned gap, our framework provides a structure that we believe can guide future developments in deep learning models requiring structured multi-dimensional composition. In the following sections, we provide background, formalize the framework and its properties, and outline potential applications in machine learning along with future research directions that we plan to pursue.

\section{Background and Related Work}
\subsection{Monoidal Structures}
A monoidal structure consists of a set $S$ equipped with an associative binary operation $\ast: S \times S \to S$, and an identity element $e \in S$ such that $e \ast a = a \ast e = a$ for all $a \in S$. Examples include $(\mathbb{N}, +, 0)$ and $(\mathbb{R}^{n\times n}, \cdot, I)$. Monoidal categories generalize these to categorical objects and morphisms, foundational in algebra and computer science \cite{MacLane1998}.

\subsection{Bimonoidal Structures and Semirings}
Bimonoidal structures involve two operations $\oplus$ and $\otimes$ under the interchange law: $(a \otimes b) \oplus (c \otimes d) = (a \oplus c) \otimes (b \oplus d)$. Unlike semirings, they do not require distributivity, allowing more flexible modeling \cite{Aguiar2010, Bridgeman2017}.

\subsection{Multi-Monoidal and Categorical Generalizations}
N-fold monoidal categories generalize monoidal structures to allow multiple independent monoidal operations in one setting \cite{Balteanu2003}. This concept permits multi-dimensional composition to be analyzed categorically. Similarly, tensor network formalisms exploit such multi-monoidal ideas to facilitate multi-dimensional algebraic computations \cite{Cichocki2016}.

\subsection{Structured Embeddings and Symbolic Composition}
For example, rotary and relative positional encodings integrate position information via fixed or learned transformations in Transformer models \cite{Su2021, Heo2024, He2020, Neelakantan2022}. However, these methods do not explicitly enforce a compositional algebra. Our framework can be seen as a generalization that encodes the history of transformations through operator composition, thereby unifying such positional embeddings under a common algebraic structure.

\section{Algebraic Framework}

\subsection{One-Dimensional Case}
Each element is a tuple $(\mathbf{a}, A)$, where $\mathbf{a} \in \mathbb{R}^n$ and $A \in \mathrm{GL}(n)$. We define:
\[
(\mathbf{a}, A) \circ (\mathbf{b}, B) := (\mathbf{a} + A \mathbf{b}, AB)
\]
This is associative but not commutative in general.

\subsection{Multi-Axis Structure}
An element is given by:
\[
\mathbf{x} = (\mathbf{a}, R_1^{n_1}, R_2^{n_2}, \dots, R_D^{n_D})
\]
with $R_i \in \mathrm{GL}(n)$ and $n_i \in \mathbb{Z}$ for each axis $i$.

Axis-specific composition is defined when all other axes match:
\[
n_j^{(x)} = n_j^{(y)} \quad \forall j \ne k
\]
Composition along axis $k$ is:
\[
\mathbf{x} \circ_k \mathbf{y} = (\mathbf{a} + R_k^{n_k} \mathbf{b}, R_1^{n_1}, \dots, R_k^{n_k + m_k}, \dots, R_D^{n_D})
\]

\subsection{Interchange Law}
The global interchange law ensures consistency:
\[
(\mathbf{x} \circ_i \mathbf{y}) \circ_j (\mathbf{z} \circ_i \mathbf{w}) = (\mathbf{x} \circ_j \mathbf{z}) \circ_i (\mathbf{y} \circ_j \mathbf{w})
\]
This holds if and only if $R_i R_j = R_j R_i$ for all $i \ne j$.

\section{Extended Algebraic Constructions}
\begin{itemize}
    \item \textbf{Identity and inverse elements per axis:} Each axis has an identity 
    element (e.g. $(0, I)$) and, if applicable, an inverse. For example, in the 
    one-dimensional case $(a, A)^{-1} = (A^{-1}(-a), A^{-1})$.
    \item \textbf{Graded embeddings with axis-specific groups:} Different axes may use 
    different transformation groups (grades), allowing heterogeneous composition 
    rules within one embedding.
\end{itemize}

\section{Compositional Representations in Machine Learning}
\subsection{Compositional Representations Across Domains}

Compositionality -- the principle that complex representations can be built by combining simpler parts -- is a fundamental property in both mathematics and machine learning. In category theory, this idea is formally captured by the composition of morphisms and monoidal structures that govern how systems combine \cite{MacLane1998}. In modern ML domains, compositional representations are critical. For instance, in natural language, the meaning of a sentence is determined by composing the meanings of its words and sub-phrases \cite{MitchellLapata2010}. In computer vision, models interpret a scene by composing representations of objects and their parts (e.g. part-whole relationships in an image).  
Graph neural networks likewise rely on composing information over nodes and edges to yield coherent graph-level representations.  
These examples illustrate that the ability to systematically combine constituent representations is essential for understanding complex structures.

\subsection{Monoidal Structures and Category Theory}
Mathematically, \textit{monoidal structures} provide a principled foundation for modeling composition. A monoid consists of an associative binary operation with an identity element, offering a formal way to combine elements repeatedly. Category theory generalizes this: categories are built on compositionality, and \emph{monoidal categories} introduce a tensor-product (parallel composition) alongside sequential composition.

\subsection{Categorical Models of Meaning}
This algebraic perspective has been applied to machine learning and semantics. For example, categorical models of meaning use grammatical structure as a guide for vector composition.  
In the work of Coecke et al. (2010), a sentence's meaning is computed via morphism composition in a compact closed category, combining word vectors according to grammar-based tensors \cite{Coecke2010}. Such frameworks demonstrate how monoidal algebra (associativity, identities, and commutation constraints) can enforce consistency in composed representations.

\subsection{Matrix-Based Sequence Encodings}
Beyond theoretical frameworks, several machine learning models have directly incorporated algebraic composition operations. In natural language processing, one line of work represents words or phrases as matrices (or matrix-vector pairs) that act as linear transformers on other word embeddings. For example, adjectives have been modeled as matrices that modify the vectors of nouns they accompany \cite{BaroniZamparelli2010}. This idea was extended in recursive neural networks, where each word is associated with a vector and a matrix, and composition is performed by matrix multiplication and vector addition \cite{Socher2012}. These \emph{matrix-based sequence encodings} enable recursive or hierarchical models to capture context-sensitive meaning beyond simple commutative operations.

\subsection{Non-Commutative Composition in Sequence Models}
Rudolph and Giesbrecht (2010) introduced a compositional matrix-space model treating semantic composition as non-commutative matrix multiplication \cite{Rudolph2010}. Such approaches show that non-commutative algebra (e.g. matrix multiplication, which does not swap order) can effectively model the importance of word order and structural roles in sequences. In contemporary architectures like the Transformer, composition is implicit: attention mechanisms mix token representations, and positional encodings are added to inject sequence order \cite{Vaswani2017}. However, these methods do not enforce an explicit algebraic law of composition; the operations (addition of position vectors or learned attention patterns) are chosen for convenience rather than derived from first principles.

\subsection{Structured State Space Models}

Structured State Space Models (SSMs)—such as S4 \cite{gu2022}, Hyena \cite{poli2023}, Mamba \cite{gu2023}, and S4ND \cite{mehta2023}—offer a compelling alternative to attention mechanisms by modeling sequences through linear dynamical systems with learned transition structures. These models update tokens via structured scans, where each step evolves according to recurrence relations defined by transition and projection matrices. A key advantage of SSMs lies in their inductive bias toward continuous-time dynamics, which naturally supports efficient modeling of long-range dependencies in data with temporal or spatial continuity—such as audio, video, or time series. This structured bias enables smooth information propagation and scale-invariant representations.

Our framework generalizes this: it extends the algebraic principles underlying SSMs, inherits their favorable properties, and offers a complementary perspective that broadens the understanding and applicability of such models.

\subsection{Motivation for a Directional Non-Commutative Framework}
This gap motivates our framework: by using a directional non-commutative monoidal structure, we obtain well-defined composition operators with associative and interchange properties, providing a rigorous yet flexible way to encode sequences, images, and other structured data in a compositional manner. Our approach leverages the strengths of prior algebraic methods (e.g. matrix-based composition) and categorical insights, while extending them to multiple axes with a coherent global law.

\section{Potential Applications in Machine Learning and Future Research}
We now discuss potential applications of our proposed algebraic structure across several machine learning domains and outline promising directions for future research. These avenues remain to be implemented and validated, but they illustrate how the new structure could meaningfully influence the design of deep learning models. We plan to pursue this line of investigation in our own future work.

\subsection{Compositional Embeddings Across Modalities}
% From \subsection{Embedding Composition}
In 1-D, each element is represented as a tuple $e_i = (\mathbf{v}_i, R_i)$. For a sequence $x_1, x_2, \ldots, x_T$, we compose the element embeddings sequentially to encode the entire sequence:
\[
E = (((e_1 \circ e_2) \circ e_3) \dots \circ e_T) = \mathbf{v_1} + R_1 \mathbf{v_2} + R_2 R_1 \mathbf{v_3} + \ldots + R_{T-1} R_{T-2} \ldots R_2 R_1 \mathbf{v_T}
\]

% From \subsection{Image Embeddings}
In 2-D, an element can be written as $e_{ij} = (v_{ij}, R_x, R_y)$ for positions $1 \le i \le H$ and $1 \le j \le W$. Composing all elements of an $H \times W$ grid yields:
\[
E = \bigcirc_{i=1}^H \bigcirc_{j=1}^W e_{ij} = \sum_{i=1}^H \sum_{j=1}^W R_x^{i-1} R_y^{j-1} v_{ij}
\]

Here we use two composition operators, $\circ_x$ and $\circ_y$, corresponding to moves in the horizontal ($x$) and vertical ($y$) directions, respectively. Each pixel or cell has associated transforms $R_x$ and $R_y$ for motion along those axes. Composition along the $x$-axis accumulates transformations across columns, and along the $y$-axis accumulates transformations across rows. The result is a learnable two-dimensional positional encoding that preserves ordering in both dimensions. 

Analogously, many signals (e.g., audio spectrograms or EEG plots) are naturally arranged on a time–frequency grid. We can assign one composition operator $\circ_t$ for the temporal axis and another $\circ_f$ for the frequency axis. This yields a structured encoding of positions in time–frequency representations, enabling models such as 2D Transformers to incorporate both temporal and spectral locality in a principled way. 

We can further extend to 3D data such as video. A video can be viewed as an $H \times W \times T$ grid (two spatial dimensions plus time). By introducing a temporal composition operator $\circ_{\text{time}}$ in addition to the spatial ones, we obtain a 3-axis embedding for video data. For example, a single frame can be represented as $(v; R_{\text{row}}, R_{\text{col}}, R_{\text{time}})$. Composing frames along the time axis accumulates transformations that capture dynamics across frames. Thanks to the interchange law, composing spatially within each frame and then advancing in time yields the same result as first advancing each pixel in time and then composing spatially. This coherence is valuable for video models, as it ensures consistent representations of motion regardless of composition order.

\subsection{Shift-Invariant Multi-Axis Representations}

A central benefit of the multi-axis algebra is the ability to derive representations that are invariant to translations (shifts) along one or more axes, without averaging away the internal structure. We now present the concept of an \emph{$m$-representation}, a compositional embedding that achieves this balance of \emph{local order-sensitivity} and \emph{global shift-invariance}. We first describe the construction for one-dimensional sequences, then generalize to multiple dimensions.

Consider a sequence of length $N$ with element embeddings $a_1, a_2, \dots, a_N$, where each $a_i \in \mathbb{R}^d$. Choose a window length $m$ ($1 \le m \le N$). Slide a window of size $m$ across the sequence and aggregate contributions from each position to form the representation.

For each window starting at position $k$ ($k=1,2,\dots,N-m+1$), we define a \textbf{window embedding} $s_k$ by applying position-dependent rotary transformations to the window’s elements and summing them:
\begin{equation}
\label{eq:window-sum}
    s_k = \sum_{i=1}^{m} R^{\,i-1} a_{\,k+i-1}\,,
\end{equation}
where $R$ is a fixed block-rotation matrix of dimension $d$ where each block is $K\times K$ and $d=mK$. Here $R^{\,i-1}$ applies $R$ $(i-1)$ times, so $s_k = a_k + R\,a_{k+1} + \cdots + R^{m-1} a_{k+m-1}$ encodes the local order of elements. Different orderings of the $a_i$ produce different $s_k$.

Next, partition $s_k\in\mathbb{R}^{mK}$ into $K$ blocks $s_{k,j}\in\mathbb{R}^m$ (Note that the $j^{th}$ $K \times K$ block in $R$ acts solely on $s_{k, j}$), and define the \textbf{magnitude vector} $v_k\in\mathbb{R}^K$ by:
\begin{equation}
    v_k = (\|s_{k,1}\|,\;\|s_{k,2}\|,\;\dots,\;\|s_{k,K}\|).
\end{equation}
This discards phase information and captures block magnitudes. Summing these $v_k$ yields:
\begin{equation}
    v = \sum_{k=1}^{N-m+1} v_k\,,
\end{equation}
giving the $m$-representation $v$ for the sequence.

We generalize to $n$-dimensional signals: for an $N_1\times \cdots \times N_n$ tensor and window shape $m_1 \times \cdots \times m_n$, we form each window embedding:
\begin{equation}
\label{eq:nd-window}
    s_{k_1,\dots,k_n} = \sum_{i_1=1}^{m_1}\cdots \sum_{i_n=1}^{m_n} 
    \bigl(R_1^{\,i_1-1}\cdots R_n^{\,i_n-1}\bigr)\;
    a(k_1+i_1-1,\dots,k_n+i_n-1)\,,
\end{equation}
where $R_1,\dots,R_n$ are commuting rotation transforms for each axis. Partitioning each $s_{k_1,\dots,k_n}$ into blocks and summing their norms yields vectors $v_{k_1,\dots,k_n}$. Summing these over all window positions gives the final $m$-representation $v$ for the tensor.

The multidimensional $m$-representation retains the same traits: it captures local structure within each window and is invariant to global shifts along any axis. In practical terms, signals (e.g.\ images) that differ only by a shift produce identical representations $v$, facilitating tasks like content-based matching. We have therefore achieved a representation that is \emph{invariant} to translations while preserving local order.

\subsection{Axis-$k$ Transform for Alignment in Multidimensional Signals}

Consider two structured embeddings $X$ and $Y$ that we wish to align along axis $k$. Suppose $X$ has length $n_k$ along axis $k$. We define the \textbf{axis-$k$ transform} $R_k$ as a shift operator along axis $k$: $R_k^s Y$ shifts $Y$ by $s$ units on that axis. For example, $R_k^{n_k}Y$ concatenates $Y$ immediately after $X$ along axis $k$.

More generally, one can search for the shift $s$ that best overlaps $Y$ with $X$. For instance, choosing 
\[ 
s^* = \arg\max_{s} \langle X,\;R_k^s Y\rangle 
\] 
finds the optimal alignment offset $s^*$ along axis $k$.

This idea extends naturally to $d$ dimensions. Let $R_1,\dots,R_d$ be shift transforms for each axis. To align $Y$ with $X$ in all axes, apply shifts $s_1,\dots,s_d$ on each axis: 
\[ 
Y' = R_1^{s_1}\cdots R_d^{s_d} Y. 
\] 
Because the $R_i$ commute, the order of shifts does not matter. In summary, powers of $R_k$ let us reposition and align entire structures. This enables operations like sequence alignment or embedding fusion by sliding one structure relative to another.

\subsection{Concatenation of Structured Embeddings}

We now describe how to combine two structured embeddings $X$ and $Y$ along a specified axis $k$. 
Suppose
\[
    X = (a; R_1^{n_1},\dots,R_k^{n_k},\dots,R_D^{n_D}), \qquad
    Y = (b; R_1^{m_1},\dots,R_k^{m_k},\dots,R_D^{m_D}),
\]
where $a,b\in\mathbb{R}^d$ are the content vectors and $R_i^{n_i},R_i^{m_i}$ encode each embedding's extent. Assuming other axes are compatible in size, we want $X \oplus_k Y$, the concatenation along axis $k$.

To align $Y$ after $X$ on axis $k$, we use the axis-$k$ shift: multiply $Y$'s vector by $R_k^{n_k}$. In a simple concatenation, this shift is $s=n_k$ so that $Y$ starts immediately after $X$.

We define:
\begin{equation}
\label{eq:concat}
    X \oplus_k Y = \bigl(a + R_k^{n_k} b \;;\; R_1^{u_1}, \dots, R_k^{u_k}, \dots, R_D^{u_D}\bigr),
\end{equation}
where $u_k = n_k + m_k$ and $u_i = \max(n_i,m_i)$ for $i\neq k$. Here $a + R_k^{n_k} b$ adds $Y$'s content shifted by $n_k$ units to $X$'s content.

This $\oplus_k$ operation remains associative and reversible under our framework. For example, concatenating two images $X=(a;R_x^{N},R_y^{H})$ and $Y=(b;R_x^{M},R_y^{H})$ along $x$ gives
\[
X \oplus_x Y = (a + R_x^{N} b \;;\; R_x^{N+M}, R_y^{H}),
\]
placing $Y$ to the right of $X$ with width $N+M$.

In summary, $\oplus_k$ aligns and merges two embeddings along axis $k$ by shifting one by $R_k^{n_k}$ and adding vectors. It preserves alignment information and fits naturally into our algebraic framework.

\subsection{Non-Commutative Transformer Self-Attention}

We can encode relative positions in self-attention via our non-commutative transforms. For a sequence of length $T$, represent position $i$ by $(v_i,R_i)$ where $v_i$ is its content vector and $R_i$ is a positional transform. Define the relative transform
\[
T_{p,q} = R_q\,R_{q+1}\cdots R_{p-1},\qquad (T_{q,p}=T_{p,q}^{-1},\;T_{p,p}=I).
\]
where $q < p$. This $T_{p,q}$ denotes the "journey" from $q$ to $p$. We then apply it to keys and values:
\[
    \tilde{K}_{p,q} = T_{p,q}K_q,\qquad \tilde{V}_{p,q} = T_{p,q}V_q\,.
\]
The attention weight is computed as usual:
\[
    \alpha_{p,q} = \frac{\exp(Q_p \cdot \tilde{K}_{p,q} / \sqrt{d})}{\sum_r \exp(Q_p \cdot \tilde{K}_{p,r} / \sqrt{d})}\,,
\]
and the output is $O_p = \sum_q \alpha_{p,q}\tilde{V}_{p,q}$. This defines a non-abelian self-attention: position enters multiplicatively via $T_{p,q}$.

If we let the output $O_p = \sum_q \alpha_{p,q}V_{p,q}$ and $R_i=R$ (a fixed matrix), then $T_{p,q} = R^{\, (q-p)}$ depends only on distance, and we recover the vanilla transformer with standard relative encodings such as Rotary Position Embeddings \cite{Su2021}. In general, each $R_i$ can be learned, allowing flexible, context-dependent positional biases while keeping attention dependent only on relative position.

We extend the attention mechanism to data with multiple positional axes. Let each data point $p$ have coordinates $(n_{p,1}, n_{p,2}, \dots, n_{p,D})$ along $D$ axes, with embedding $(v_p; R_1, R_2, \dots, R_D)$. For any two such points $p$ and $q$, we define the relative transform $T_{p,q}$ to simultaneously account for the "journey" from $q$ to $p$ along all axes:
\[
T_{p,q} = \prod_{i=1}^D R_i^{\, (n_{p,i} - n_{q,i})}\,.
\]
We then apply $T_{p,q}$ to the key and value associated with $q$:
\[
\tilde{K}_{p,q} = T_{p,q} K_q, \qquad \tilde{V}_{p,q} = T_{p,q} V_q\,.
\]
The attention weights and output are computed as:
\[
\alpha_{p,q} = \frac{\exp(Q_p \cdot \tilde{K}_{p,q}/\sqrt{d})}{\sum_r \exp(Q_p \cdot \tilde{K}_{p,r}/\sqrt{d})}, \qquad O_p = \sum_q \alpha_{p,q} \, \tilde{V}_{p,q}\,.
\]
This is the multi-dimensional generalization of the non-commutative attention mechanism. Each axis-specific difference contributes a factor to $T_{p,q}$, and because all $R_i$ commute, the order of applying these shifts does not matter. The attention computation now depends on the full $D$-axis offset between $p$ and $q$, allowing the model to attend to others based on multi-axis relative positions (for example, considering both horizontal and vertical offsets in an image, or temporal and spatial differences in a video) in a unified manner.

Building on this mechanism, we can design Transformer architectures that use these compositional embeddings. Each input element is associated with directional transforms along each axis, and composition is performed along structured axes. Attention is guided by the journey-based operator $T_{p,q}$, which encodes the path from one element to another. For instance, in the case of 1D, this allows the model to compose neighboring tokens in a way that respects local context while preserving global coherence—since $T_{p,q}$ is the same regardless of whether the journey traverses a single token or a block of composed tokens in a single step. The same approach naturally extends beyond sequences to higher-dimensional data: for instance, in images or videos, spatial and temporal axes are composed independently yet coherently, enabling the model to integrate local structure into unified, context-aware representations. 

\paragraph{Rotary Position Embeddings as a Special Case}

Consider the one-dimensional case where all positional transforms $R_i$ are tied to a single fixed matrix $R \in \mathrm{GL}(n)$. If we choose $R$ to be an orthonormal block-diagonal rotation matrix (partitioning $\mathbb{R}^n$ into 2D subspaces), then an element at position $p$ can be written as $(\mathbf{v}_p, R)$. In this setting, the \emph{relative} transform from position $q$ to $p$ is $T_{p,q} = R^{\, (p-q)}$. When applied inside a self-attention mechanism (e.g., rotating key and value vectors by $T_{p,q}$ before dot-product), this ensures that attention depends only on relative position. This exactly recovers the effect of \emph{Rotary Position Embedding} (RoPE), wherein multiplying queries and keys by position-dependent rotation matrices makes their inner product a function only of relative position~\cite{Su2021}.

Extending this idea to images (two-dimensional grids) is straightforward under our multi-axis framework. We assign two base rotation matrices, $R_x$ and $R_y$, to serve as the transforms for the horizontal ($x$) and vertical ($y$) axes, respectively. To encode a position $(i,j)$ in an $H \times W$ image, we represent the element as $(\mathbf{v}_{ij};\,R_x,\,R_y)$ with $R_x R_y = R_y R_x$, satisfying the interchange law by construction. In this setting, the relative transform is $R_x^{\,(i-i')} R_y^{\,(j-j')}$. This precisely matches the 2D RoPE scheme proposed for Vision Transformers~\cite{Heo2024}, where rotary embeddings are applied independently along each axis. In practice, the embedding dimensions are split between the two axes, and the 1D RoPE rotation is applied separately to each half—one for the row index $i$, one for the column index $j$. This demonstrates that \emph{Rotary Position Embedding}, both in its original 1D form and its 2D extension, emerges as a special case of our compositional embedding framework.

\subsection{Compositional Transformers and Non-Abelian Attention Mechanisms 
Generalize both Structured State Space Models and Transformers}
  
Notably, compositional transformers and non-commutative self-attention provide a unified generalization of sequence modeling paradigms such as structured state-space models and transformer self-attention. In particular, if the attention weights are made implicit (uniform) and the token interactions are constrained to a learned recurrent transition structure, our framework reduces to a structured state-space model. Conversely, assuming a trivial flat structure over tokens (no explicit recurrence beyond sequential order) and using the standard dot-product mechanism for attention, we recover the familiar transformer self-attention architecture. Thus, both the recurrence-centric SSM and the attention-centric transformer emerge as special cases of the proposed approach, underscoring its ability to capture implicit structured recurrence as well as explicit dynamic attention. For example, consider the following expressions. For SSM:
\begin{equation}
y_k = \sum_{i \leq k} C_k \left( \prod_{j=i}^{k-1} A_j \right) B_i x_i
\end{equation}
For vanilla transformers:
\begin{equation}
y_k = \sum_{i \leq k} \alpha_{ik} V_i
\end{equation}
For compositional transformers shown here:
\begin{equation}
y_k = \sum_{i \leq k} \alpha_{ik} \left( \prod_{j=i}^{k-1} R_j \right) V_i
\end{equation}

Note how compositional transformers generalize both SSM and vanilla transformers.

\section{Computational Efficiency and Parallel Structure}

A notable advantage of the proposed framework is its compatibility with modern developments in efficient sequence modeling. The compositional operations are inherently associative, enabling flexible grouping and parallel evaluation of intermediate terms. This associativity, combined with the recursive structure of the update rule, allows for the use of efficient scan algorithms and blockwise parallelism. In particular, when the structural operators \( R \) are instantiated as block-diagonal matrices composed of \( 2 \times 2 \) rotation matrices, the resulting updates are not only expressive but also computationally lightweight. These structured transformations can be implemented with minimal overhead using hardware-efficient routines such as vectorized trigonometric kernels or fused rotation primitives. As a result, the framework integrates seamlessly with optimizations developed for both transformers and state space models, including low-rank compression, GPU-accelerated batching, and memory-efficient attention variants—making it well-suited for high-performance training and deployment.

\section{Limitation}
A key limitation of the proposed framework is the absence of experimental validation, which leaves questions open regarding its stability, susceptibility to design violations, and potential computational overhead—particularly due to the per-axis transformation operators \( R_i \in \mathrm{GL}(n) \), which may become expensive to compute and store in high-dimensional settings. A practical remedy is to adopt an efficient parameterization of each \( R_i \) as a block-diagonal matrix composed of independent \( 2 \times 2 \) rotation matrices. This choice offers several advantages: first, it ensures numerical stability and norm preservation due to the orthogonality of rotations; second, it significantly reduces computational cost, as the product of two such rotations corresponds to the \emph{simple addition of angles}, enabling compositions to be computed by summing angle parameters and reconstructing rotation matrices via trigonometric functions; and third, it guarantees that all \( R_i \) commute pairwise, since rotations in 2D subspaces naturally commute. This property ensures that the interchange law holds by construction, providing a principled foundation for consistent multi-axis composition. Moreover, the use of such structured transformations introduces a strong inductive bias, which reduces the number of learnable parameters and helps maintain model complexity at a manageable scale—an important consideration for scalable and efficient deployment.

\section{Conclusion}

We have introduced a novel algebraic framework that enables recursive, structured embeddings through axis-specific non-commutative composition operators, governed by a globally consistent interchange law. This construction unifies a wide range of modeling paradigms—including structured state-space models, Transformer self-attention, and compositional embedding methods—under a single, principled system for multi-dimensional composition.

By assigning distinct operators to each axis and enforcing consistency across dimensions, our framework offers a mathematically grounded approach to modeling complex compositional structures in data. This opens the door to new classes of architectures capable of structured attention, symbolic reasoning, and logic-inspired composition within neural systems.

Crucially, we demonstrate that several well-known mechanisms—such as sequence matrix encodings, SSM recursions, and Rotary Positional Embeddings (RoPE) in Transformers—emerge as special cases of our general formulation, underscoring the framework’s expressive breadth and theoretical coherence.

While this paper has focused on the theoretical foundations, we have outlined several concrete avenues for integrating our framework into existing architectures—for example, enhancing positional encodings in Transformers or generalizing recurrence mechanisms in state-space models. We plan to empirically evaluate these ideas in follow-up work aiming to demonstrate measurable gains in performance, interpretability, or architectural generality relative to current baselines.

% Placeholder for acknowledgment and patent notice
% To be included in camera-ready version:
% "We thank [Full Name] for their valuable contributions. A patent application 
% has been filed for aspects of this work."

\bibliographystyle{plain}

\end{document}